\documentclass[lettersize,journal]{IEEEtran}
\usepackage[numbers]{natbib}
\usepackage[T1]{fontenc}
\usepackage[utf8]{inputenc}
\usepackage{graphicx}
\usepackage{multirow}
\usepackage{amsmath,amssymb,amsfonts}
\usepackage{amsthm}
\usepackage{mathrsfs}
\usepackage{xcolor}
\usepackage{textcomp}
\usepackage{manyfoot}
\usepackage{booktabs}
\usepackage{algorithm}
\usepackage{algorithmicx}
\usepackage{algpseudocode}
\usepackage{listings}
\usepackage{xpatch}
\usepackage{xspace}
\usepackage{bm}
\usepackage{tabularx}
\usepackage{adjustbox}
\usepackage{multirow}
\usepackage{url}

\newtheorem{theorem}{Theorem}
\newtheorem{proposition}{Proposition}

\newtheoremstyle{named}{}{}{\itshape}{}{\bfseries}{:}{.5em}{\thmnote{#3}#1}
\theoremstyle{named}
\newtheorem*{nameddef}{}

\begin{document}
\newcommand{\uj}{\text{UJ}}
\newcommand{\ai}{\text{AI}}
\newcommand{\sd}{\text{SD}}
\markboth{IEEE Transactions on Human-Machine Systems,
Vol. X, No. X, March 2026}
{A. Campagner
\MakeLowercase{\textit{(et al.)}:
When Do Humans and Machines Truly Complement Each Other?}}

\title{When Do Humans and Machines Truly Complement Each Other? Formal Conditions and Empirical Evidence for Human-AI Synergy}

\author{Andrea Campagner, Federico Cabitza, Vincenzo Crupi \thanks{A. Campagner is with Department of Informatics, Systems and Communication, University of Milano-Bicocca, Milan, Italy and IRCCS Ospedale Galeazzi Sant'Ambrogio, Milan, Italy} \thanks{F. Cabitza is with Department of Informatics, Systems and Communication, University of Milano-Bicocca, Milan, Italy and Digital Health \& Wellbeing Center, Fondazione Bruno Kessler (FBK), Trento, Italy} \thanks{V. Crupi is with Department of Philosophy and Education Sciences, University of Turin, Torino, Italy}}




\maketitle
\begin{abstract}
Can humans and artificial intelligence cooperate such that their combination yields greater accuracy than either agent alone? This result, termed \emph{synergy} or \emph{complementarity}, is notoriously difficult to achieve, even when both agents possess significant independent accuracy. In this paper, we present a formal framework to model human-AI interaction through the lens of reliance patterns, which characterize the joint correctness structure of human and algorithmic decisions. We derive mathematically necessary and sufficient conditions for synergy, demonstrating that its emergence depends on the ability of either agent to identify and correct the errors of the other, what we call Human and Algorithmic Correction. Beyond its theoretical contribution, the proposed framework provides a foundation for the design and evaluation of knowledge-based decision support systems, where interaction policies must dynamically integrate heterogeneous sources of expertise. We demonstrate this through the analysis of seven clinical case studies. Our empirical results reveal a ``dominance trap,'' where a strong algorithmic advantage can paradoxically inhibit synergy by eroding human corrective intervention. We also show that synergy can persist even in the presence of cognitive biases, such as automation bias, provided there is functional de-correlation between human and AI errors. These results provide a formal foundation for key concepts in the human-AI systems literature and offer actionable insights for designing interaction-centric AI systems that foster complementarity.
\end{abstract}


\section{Introduction}

\IEEEPARstart{A}{rtificial} Intelligence is being increasingly advocated and employed as a tool to support judgment and decision making in an impressive variety of domains. The breakthrough of contemporary AI has revitalized and reshaped two long-standing inquiries: first, how human judgment compares with algorithmic outcomes across diverse contexts; and second, under what conditions the integration of these two assessments can reliably yield superior accuracy.

Current theoretical understanding appears to be rather strong for the first issue, but still limited for the second~\cite{bansal2019beyond,hemmer2025complementarity}. Crucially, a recent comprehensive review~\cite{Vaccaro2024} notes that we still lack a broad conceptual understanding of when human-AI teams outperform their individual components (“when a combination of humans and AI are better than either alone”). Consequently, the specific conditions under which human-AI systems can exhibit \emph{synergy}~\cite{Bao2023} (or \emph{complementarity}~\cite{hemmer2025complementarity}) are yet to be fully elucidated, especially as this condition often appears elusive in practical scenarios \cite{rieger2025highly}.

From a philosophy-of-technology perspective, recent work \cite{Ferrario2026Epistemology} has argued that the explanatory gap surrounding human–AI synergy is partly conceptual: synergy is not merely a stand-alone, post hoc indicator of predictive performance, but rather a form of defeasible evidence that a given human–AI interaction constitutes a reliable epistemic process. Complementing this view, recent contributions in the human–machine systems literature \cite{rieger2025highly} emphasize that synergy is not merely an outcome, but a design-dependent property emerging from the interaction between human and algorithmic components: differences in information, capabilities, and error structures between humans and AI—-rather than overall accuracy alone—-constitute the primary sources of complementarity and must be explicitly leveraged through appropriate interaction mechanisms. Taken together, these perspectives suggest that understanding synergy requires moving beyond purely outcome-based definitions toward a unified account that integrates  interaction design and the structure of human–AI error complementarities.


From a more technical perspective, the scientific community has only recently begun to address this issue by investigating theoretical formulations of synergy, and three main strands of research can be identified, which, however, have been relatively isolated from each other.  Most of the work in this setting has been developed in the theoretical machine learning literature and has focused on the study and formal analysis of the deferral paradigm, that is, developing AI systems that can defer their judgment to a human. One of the seminal contributions in this field is the work of Madras et al. \cite{Madras2018} who proposed and first analyzed the effects of deferral on human-AI collaboration relying on the framework of selective prediction in machine learning literature. Developing this contribution, Charusaie et al. \cite{charusaie2022sample} studied the theoretical properties of deferral learning systems trained to be complementary to humans; Straitouri et al. \cite{straitouri2023improving} similarly considered how synergy could be achieved through conformal prediction (a family of statistical techniques to create deferral AI systems); De Toni et al. \cite{detoni2024towards} studied the computational complexity of achieving complementary through deferral, showing that in general the problem is computationally infeasible.  Although this body of work characterizes conditions for synergy, it remains largely inapplicable to traditional AI systems that provide judgments without deferral. Furthermore, these studies emphasize the algorithmic perspective, leaving the more interactional dimensions of the collaboration relatively unexplored.
Crucially, the deferral paradigm focuses on a logic of \emph{delegation}, where the final judgment is produced by only one of the two agents. In contrast, our work addresses \emph{judgment integration} (or co-agency), where both human and AI assessments are available and must be combined. 

A different stream of work, more focused on the interactional aspects, has aimed at explicitly studying combination mechanisms to model the interaction between AI and human agents. In this light, the Bayesian paradigm, which models both human and AI reasoning as well as their combination as parametric probabilistic models and employs empirical interaction data to fit their parameters, has been dominant, partly due to the already rich literature on probabilistic models of human cognition~\cite{chater2006probabilistic}. Steyvers et al. \cite{steyvers2022bayesian} proposed a Bayesian framework to model and combine the predictions of human and AI agents; Westby et al. \cite{Westby2023} studied a Bayesian theory of mind framework for modeling collective intelligence (as a special form of synergy) in human-AI teams.
While these studies include a representation of the cognitive processes underlying human judgment, the framework we adopt in this work does not involve such commitment. This approach allows us to directly address the constructs relevant to human-AI interaction as they emerge from formal probabilistic constraints.
An additional limitation of Bayesian approaches is their focus on modeling joint performance rather than identifying the foundational conditions for synergy; consequently, the requirements for synergy emerge only as indirect byproducts of their parametric fits, rather than as explicit theoretical constraints. Moreover, they often rest on strong assumptions (e.g., normality, or specific constraints on combination functions), which limit their ability to yield actionable insights, particularly when such assumptions are violated. 

To address these gaps and more in line with the approach we take in this paper, a recent strand of work in the study of human-AI interaction mechanisms has aimed at characterizing conditions under which synergy can (or cannot) emerge by adopting a more general (model-free) probabilistic framework. Donahue et al. \cite{donahue2022human} studied equivalent definitions of synergy when the combination of AI and human agents is realized by linear combination rules, also providing characterizations of cases in which synergy is impossible to achieve. The study analyzed synergy solely within the constraints of linear combination rules, which assume fixed weights for each agent. Such models may fail to capture the non-linear and context-dependent interactions typical of work practices, where a human expert evaluates the quality of an AI signal on a case-by-case basis.
More recently, Peng et al. \cite{Peng2025} studied no-free lunch theorems for human-AI synergy, showing that it is generally impossible to achieve synergy if the combination rule only takes into account the overall accuracy of the human and AI agents. 
By shifting the focus from global performance metrics to the granular dynamics of reliance patterns (specifically Human Correction and Algorithmic Correction, as defined below), we demonstrate that synergy is not only theoretically possible but also empirically predictable when the interaction logic exploits and fosters the mutual reduction of errors on a case-by-case basis.


The goal of this contribution is to advance insight on these matters: to do so, we adopt a probabilistic approach, similar to those previously considered in~\cite{donahue2022human,Peng2025}, which aims at addressing the limitations of previous analytical studies on human-AI synergy. In contrast to prior research, our framework avoids parametric assumptions regarding both how agents formulate decisions and how their judgments are combined, and instead adopts a non-parametric and model-free approach. 

Thus, we make two main contributions.
\begin{itemize}
    \item First, we propose a general framework to formally define the notion of human-AI collaboration and human-AI synergy. Our framework rests on the central notion of reliance patterns~\cite{Cabitza2023}, which have been introduced in the human-AI interaction community to model and quantify issues related to AI appropriation.
    \item Second, we provide different sets of formal conditions for human-AI synergy, discussing their sufficiency and necessity. We also provide empirical illustrations of these conditions in practical settings (drawn from real-world medical scenarios of AI use).
\end{itemize}
Beyond theoretical characterization, our framework provides a set of design constraints and principles for human-AI interaction policies, enabling the engineering of synergistic systems. The conditions derived in this work can be operationalized as design objectives, evaluation metrics, and training targets for hybrid systems and we provide a discussion of the implications of our results for the design of hybrid human-AI systems.


\section{The Formal Framework}
In this section, we introduce a formal framework for modeling human–AI interaction. This framework is intentionally model-agnostic and non-parametric: rather than assuming specific cognitive or algorithmic processes, the framework captures interaction in terms of observable variables—namely, the correctness of human judgments, AI outputs, and final decisions. This abstraction enables the framework to be applied across different domains and system implementations, supporting both theoretical analysis and empirical evaluation.

Formally, let $X \in \{x,\neg x\}$ be a binary variable representing the ground-truth state of a situation of interest $S$. For instance, $x$ may denote the presence of a specific pathology in a patient’s tissue (where the patient case constitutes $S$), while $\neg x$ denotes its absence. A human user (e.g., a clinician) typically assesses the validity of $x$ against $\lnot x$ based on a complex and often heuristic array of information indicators. Let $H \in \{h,\lnot h\}$ include the possible outcomes of such human expert’s judgment process about $X$. For simplicity, we write $H=X$ to mean that either $(x\wedge h)$ or $(\neg x\wedge \neg h)$ holds. Consider also an artificial system designed to similarly make a decision about the situation of interest (e.g., in our recurring example about clinical-decision making, detecting specific pathologies) by processing fine-grained data extracted from each case. For our current purposes, we can think of an algorithm devised through machine learning to deliver an outcome in the form of a further binary variable $A \in \{a,\lnot a\}$.

Finally, when both $H$ and $A$ are available, a system (or human agent) integrates them into an overall decision $F \in \{f,\neg f\}$. We model $F$ as depending on the values of $H$ and $A$ according to some (situation-dependent) combination function:
\[
c_S : H \times A \rightarrow F.
\]

We note that, compared with previous work \cite{donahue2022human,Peng2025}, we do not make any assumption that the combination rule is solely determined by the accuracy of the human and AI agents: rather, as a consequence of the dependence of the combination rule $c$ on the situation of interest $S$, $c$ could, in principle, take into account context information that also depends on the case at hand\footnote{This more general formulation, allows the overall assessment $F$ to be obtained from $H$ and $A$ in a non-uniform manner, by accounting for situation-specific details and patterns. We model $S$ as an implicit parameter of the combination function $c$, rather than explicitly representing $c$ as a ternary function, to emphasize that $S$ has a different status from the assessments $H$ and $A$. Specifically, $S$ may depend on the particular case at hand as well as on the internal---possibly dynamic---state of the human or algorithmic agent performing the combination. Moreover, $S$ may not admit a systematic, complete, or computational description. Under this interpretation, the combination function can be understood as a policy that is conditioned on $S$ and that may be stochastic, non-deterministic. We note that such a dependence on the situation $S$ is necessary to our study: indeed, following the no-free lunch theorems in \cite{Peng2025}, synergy is generally impossible to achieve for uniform combination rules that only depend on the agents' accuracies.} (as implicitly considered, e.g., in~\cite{Cabitza2023}).

These variables define a set of 16 mutually exclusive and jointly exhaustive elementary possibilities and a related algebra on which a probability distribution $\Pr$ can be defined, as shown in the following section.

\subsection{Reliance Patterns}
Based on the agreement between $X$ and $H, A, F$, respectively, we can define so-called reliance patterns, which enumerate the 8 possible truth outcomes~\cite{Cabitza2023}. In detail, consider three additional variables, namely, the correctness of the human, the AI, and the combined system denoted as UJ (Unsupported human Judgment), AI (AI judgement), and SD (Supported final Decision), respectively. These variables are binary (0 or 1) and their values encode the correctness (i.e., agreement with $X$) of each of $H, A$ and $F$.
The 8 possible configurations are enumerated as binary reliance patterns, as detailed in Table \ref{tab:patterns}. We will also use the symbol $*$ as a shorthand for enumeration of patterns (e.g. `**1' corresponds to the configurations 001, 011, 101, 111). 

\begin{table*}[!htb]
\caption{Reliance patterns: definition of all possible decisional configurations between human decision makers and AI system. In the first three columns, 0 denotes an incorrect decision, and 1 a correct decision. We associate the attitude towards the AI in each possible reliance pattern, which leads to either accepting or discarding the AI’s advice, and the main related cognitive biases.}
\tiny
\label{tab:patterns}
\begin{tabularx}{\textwidth}{XXXXXX}
\hline 
{\textbf{Human judgment} (\uj)} & {\textbf{AI support
} (\ai)} & {\textbf{Final decision} (\sd)} & \textbf{Reliance pattern} & \textbf{Other formulations} & \textbf{Facilitating Biases} \\ \hline
wrong (0) & wrong (0) & wrong (0) & detrimental reliance  & /      & automation complacency          \\
wrong (0) & wrong (0) & right (1) & beneficial under-reliance & / & extreme algorithmic aversion    \\ 
wrong (0) & right (1) & wrong (0) & detrimental self-reliance  & Detrimental overriding   & self-anchoring bias               \\
wrong (0) & right (1) & right (1) & beneficial over-reliance & Correct adherence    & algorithm appreciation                     \\
right (1) & wrong (0) & wrong (0) & detrimental over-reliance  & Detrimental adherence  & automation bias                 \\
right (1) & wrong (0) & right (1) & beneficial self-reliance   & Corrective overriding   & algorithmic aversion            \\
right (1) & right (1) & wrong (0) & detrimental under-reliance & / & extreme algorithmic aversion    \\
right (1) & right (1) & right (1) & beneficial reliance   & /      & confirmation bias (in later cases) \\ \hline
\end{tabularx}%
\end{table*}

While, obviously, knowledge of $X, H, A, F$ is sufficient to determine $\uj, \ai, \sd,$ the converse does not hold. 
At the same time, one of $X,H,A,F$, together with $\uj, \ai, \sd$ is sufficient to determine the remaining variables. Consequently, these formulations are functionally equivalent; hereafter, we use either notation interchangeably to facilitate ease of exposition.

From a systems perspective, all the introduced variables and reliance patterns correspond to measurable quantities in human–AI experiments or deployments. For example, H, A, and F can be directly observed in decision-support workflows, while X corresponds to ground truth obtained through validation or expert consensus, which would then enable the identification of reliance patterns and the quantification of their frequency. This allows the proposed framework to serve not only as a theoretical model, but also as a basis for the instrumentation and evaluation of human–AI systems, as we will illustrate in Section \ref{sec:empirical}.

\section{The default setting}
In this section, we introduce a set of baseline assumptions (that jointly define what we call the \emph{default setting}) and definitions aimed at characterizing conditions under which a human–AI system can achieve a reasonable level of accuracy, independent of whether true synergy is attained. These assumptions are widely adopted, either explicitly or implicitly, across the literature on human–AI decision making and machine learning. As such, they provide a standard reference setting for analyzing the behavior of hybrid systems. In this direction, the notation described in the previous sections is sufficient to define three key notions concerning accuracy of human-AI hybrid systems:

\begin{nameddef}[Algorithmic Advantage (A-Adv)] $\Pr(A = X) > \Pr(H = X)$ (i.e., $\Pr(*1*) > \Pr(1**)$).
\end{nameddef}

\begin{nameddef}[Human Advantage (H-Adv)] $\Pr(H = X) > \Pr(A = X)$ (i.e., $\Pr(1**) > \Pr(*1*)$).
\end{nameddef}

\begin{nameddef}[Synergy]
$$\text{S1}) \quad \Pr(F=X) > \Pr(A = X) \text{ (i.e., } \Pr(**1) > \Pr(*1*)),$$
$$\text{S2}) \quad \Pr(F=X) > \Pr(H = X) \text{ (i.e., } \Pr(**1) > \Pr(1**)).$$
\end{nameddef}

Intuitively, Algorithmic Advantage formalizes the frequent empirical observation that AI systems can outperform humans in decision-making tasks. Similarly, Human Advantage formalizes the opposite situation in which humans outperform AI systems, which is still typical for particularly complex tasks that elude pattern recognition capabilities.

Furthermore, \textit{Synergy}, as already described intuitively in the Introduction, formalizes the primary objective of human-AI interaction: the emergence of a hybrid system more accurate than either agent in isolation~\cite{rieger2025highly}. It is easy to see that condition S2 corresponds to a form of \textit{Human Augmentation}, by which the interaction between the human and AI agents enhables the system to reach a performance superior to that of the human alone. By symmetry, condition S1 represents a form \textit{AI Augmentation}, that is the requirement that the hybrid human-AI system outperforms its purely algorithmic component. 

In relevant applications, significant features of the probability distribution $\Pr$ as referred to a specific domain and target case(s) will have to be constrained by reliable empirical input. In our theoretical analysis in this section, however, we will only put forward a small set of general assumptions about $\Pr$.

\begin{nameddef}[Conditional Independence (CI)]
$H \perp_{\Pr} A \,|\, X.$
\end{nameddef}
This assumption is implicitly given by the graphical model in~\cite{Cabitza2023} and corresponds to an idealized setting in which neither the human nor the algorithm has access to the other’s output at the time of producing $H$ and $A$, respectively. Although this may not hold in many real deployments, it provides a tractable and realistic baseline.

\begin{nameddef}[Human Reliability (HR)]
$0.5 < \Pr(h|x), \Pr(\lnot h|\lnot x) < 1.$
\end{nameddef}

HR characterizes human judgment as valuable, although imperfect. It implies:
\[
\Pr(h \land x) > \Pr(\lnot h \land x) > 0, \quad 
\Pr(\lnot h \land \lnot x) > \Pr(h \land \lnot x) > 0,
\]
and, denoting the correctness condition as $H = X$:
\begin{equation}
0.5 < \Pr(H = X) = \Pr(1**) < 1.\nonumber
\end{equation}

\begin{nameddef}[Algorithmic Reliability (AR)]
$0.5 < \Pr(a|x), \Pr(\lnot a|\lnot x) < 1.$
\end{nameddef}
This is the analog of HR for the AI system, implying:
\begin{equation}
0.5 < \Pr(A = X) = \Pr(*1*) < 1.\nonumber
\end{equation}

While the above three assumptions are considered standard in the human-AI interaction literature, other, stronger, requirements have been posited. As an example, the following requirement originally proposed by Meehl \cite{Meehl1954}, can be easily seen to be a stronger version of Algorithmic Advantage:

\begin{nameddef}[Meehl’s Postulate (MP)]
$\Pr(x \land a) > \Pr(x \land h), \quad \Pr(\lnot x \land \lnot a) > \Pr(\lnot x \land \lnot h).$
\end{nameddef}
Meehl's Postulate goes beyond Algorithmic Advantage by asking not merely that the AI is more accurate (on average) than the human, but rather that the algorithm outperforms the human in both detecting and rejecting the disease. In the same contribution, Meehl also discussed the following assumption:

\begin{nameddef}[Broken Leg (BL)]
$\Pr(x \land \lnot a \land h) > 0, \quad \Pr(\lnot x \land a \land \lnot h) > 0.$
\end{nameddef}

These highlight relatively uncommon “broken leg” cases \cite{Meehl1954}, in which human judgment correctly identifies algorithmic errors, and can be expressed equivalently as:
\[
\Pr(10*) > 0, \quad \Pr(01*) > 0.
\]


\subsection{Combination rules and baseline performance}

We now examine simple rules that may be adopted to combine human and algorithmic judgments, and contrast their accuracy with each component in isolation. The first such rule we examine corresponds actually to a family of rules satisfying a common structural constraint requiring that their output may differ only in cases of disagreement between human and AI system. As we will see in the following, this requirement on combination rules will be crucial in the definition of conditions that allow the emergence of human-AI synergy.

\begin{nameddef}[Agreement (Agr)] If $A = H$, then $F = A = H$.
\end{nameddef}
Equivalently, Agreement can be expressed as the structural constraint that $\Pr(001) = \Pr(110) = 0$, as the involved cases are those in which $A, H$ coincide, but $F$ differs from both of them.
A simple consequence of adopting a function $c$ that uniformly satisfies agreement is the following:
\begin{proposition}
    Assume the default setting. If $c$ satisfies \text{Agr} then $\Pr(F = X) > 0.25$.
\end{proposition}

While agreement alone is insufficient to guarantee $\Pr(F = X) > 0.5$, each of the following uniform combination rules can do so under our current assumptions:\footnote{Specifically, AA and FA always ensure that $\Pr(F=X) > 0.5$, DS and DA only do so under a technical condition on the prevalence.}

\begin{nameddef}[Algorithmic Aversion (AA)] $F = f$ iff $H = h$.\end{nameddef}

\begin{nameddef}[Full Automation (FA)] $F = f$ iff $A = a$.\end{nameddef}

\begin{nameddef}[Diagnostic Skepticism (DS)] $F = f$ iff $H = h \land A = a$.\end{nameddef}

\begin{nameddef}[Diagnostic Alertness (DA)] $F = f$ iff $H = h \lor A = a$.\end{nameddef}

The first two rules represent extreme reliance strategies: total dependence on either the human expert or the algorithm. DS mandates consensus between both sources to confirm a decision, whereas DA treats either source's positive assessment as sufficient.
Under the default setting and MP, FA is expected to outperform AA in terms of accuracy. DS may outperform AA when prevalence is low (reducing false alarms), while DA may outperform FA when prevalence is high. These simple rules show how distributional features affect the relative performance of combination strategies, and how Synergy requires more structured approaches than mere automation.

\section{Beyond automation: a condition for synergistic interaction}
While the conditions studied in the previous section are sufficient to ensure that simple combination rules achieve non-trivial levels of accuracy, they do not, in general, guarantee that the resulting human–AI system outperforms its individual components.
In this section, to address this gap, we move beyond this default setting to identify specific conditions under which synergy can emerge. In particular, we shift the focus from general accuracy properties to the structure of disagreements and error correction mechanisms between human and AI agents. This perspective reveals that synergy is not a direct consequence of individual accuracy, but rather an emergent property of how conflicting judgments, that are, the cases in which a combination rule satisfying Agreement can intervene, are resolved within the interaction process. The core premise is that synergy emerges only when disagreements are predominantly resolved in favor of the correct actor, regardless of which source it is.  Inspired by this observation, we introduce the two following definitions:

\begin{nameddef}[Human Correction (HC)]
$\Pr(A \neq X | F = H \neq A) > \frac{1}{2}.$
\end{nameddef}
\begin{nameddef}[Algorithmic Correction (AC)]
$\Pr(H \neq X | F = A \neq H) > \frac{1}{2}.$
\end{nameddef}

The two assumptions can be equivalently expressed as:
\begin{align}
&Pr(101) > Pr(010) \quad [HC]\nonumber\\
&Pr(011) > Pr(100) \quad [AC]\nonumber
\end{align}

Intuitively, HC means that whenever the final decision $F$ follows the human judgment against the algorithm, it is more likely that the algorithm is wrong than right. Intuitively, this condition captures the idea that the human is able to identify “broken leg” cases more often than fabricating exceptions. Symmetrically, AC means that when the final decision $F$ follows the algorithm, it is able to correct human errors. Thus, both assumptions can be understood as formalizing the requirement that the undesirable situation in which one of the two agents \emph{drags down} the performance of the entire system \cite{rieger2025highly} are \emph{sufficiently rare}. Given these assumption, we now prove our main results concerning Synergy.

\begin{theorem}
\label{th:hcacasyn}
    Assuming Human and Algorithmic Correction, the Agreement property becomes a sufficient condition for Synergy. Similarly, assuming Agreement, Human and Algorithmic Correction are sufficient for Synergy. Conversely, Synergy and Agreement imply Human and Algorithmic Correction
\end{theorem}


As already hinted at previously, Theorem \ref{th:hcacasyn} provides a key insight for the design of human–AI systems: synergy can emerge only when disagreements between human and AI are resolved in favor of the correct agent. This implies that the effectiveness of such systems depends less on the individual accuracy of their components, and more on the quality of the interaction policy governing how disagreements are handled.

In fact, part of the previous result can be strengthened by introducing a new assumption which relaxes Agreement: rather than requiring that combination rules only really operate in cases of disagreement, we weaken this requirement by only assuming that they ensure that cases of negative algorithmic aversion (i.e., cases of the form 110) are rarer than positive ones (i.e., cases of the form 001):

\begin{nameddef}[Beneficial Aversion (BA)]
$Pr(F=X|H=A\neq F) \geq \frac{1}{2},$
\end{nameddef}
which can equivalently be expressed as:

$$
Pr(001) \geq Pr(110)
$$


Based on this alternative condition, we can prove a second characterization of Synergy:
\begin{theorem}
    Assume Human Correction and Algorithmic Correction. Then Beneficial Aversion is sufficient for Synergy. The converse part of Theorem \ref{th:hcacasyn}, however, does not hold.
\end{theorem}

    
The previous results illustrate that the ability of the combination rule to arbitrate disagreements in favor of the correct agent, as formalized through the Human and Algorithmic Correction assumptions, is essentially sufficient and necessary for synergy (assuming that the combination rule itself satisfies Agreement). However, these conditions may be difficult to enforce directly in practice. In many real-world settings, by contrast, it is more straightforward to determine whether one agent systematically outperforms the other. The following propositions show that, under such performance asymmetries, the characterization of synergy can be significantly simplified. In particular, when one agent is globally more accurate than the other, synergy reduces to the presence of a single corrective mechanism (Human Correction in the case of Algorithmic Advantage, and Algorithmic Correction in the case of Human Advantage) while the other automatically emerges.

\begin{proposition}[HC $\Leftrightarrow$ Synergy under A-Adv and Agreement]
\label{prop:hcsmpa}
Assuming Algorithmic Advantage and Agreement, Human Correction is necessary and sufficient for Synergy. It is sufficient also if Agreement is replaced by Beneficial Aversion.
\end{proposition}


\begin{proposition}[AC $\Leftrightarrow$ Synergy under H-Adv and Agreement]
\label{thm:main}
Assuming Human Advantage and Agreement, Algorithmic Correction is necessary and sufficient for Synergy. It is sufficient also if Agreement is replaced by Beneficial Aversion.
\end{proposition}

\subsection{Appropriate Reliance and Synergy: an Alternative Characterization}
While in the previous section we characterized synergy in terms of probabilistic correction mechanisms, the human–machine systems literature has often approached the problem from a complementary perspective, focusing on observable patterns of user reliance on AI systems and related human factors \cite{chen2014human}. In particular, the notion of appropriate reliance~\cite{Lee2004,Schemmer2023} (AR) captures the extent to which users follow AI recommendations when they are correct and disregard them when they are incorrect: a behavior widely regarded as a key indicator of effective human–AI interaction. Bridging these perspectives is important for both theoretical and practical reasons. On the one hand, appropriate reliance provides a behavioral and empirically measurable construct, grounded in user interaction data. On the other hand, it offers a potential operational proxy for the more abstract conditions of Human and Algorithmic Correction introduced above. In this section, we therefore investigate how appropriate reliance relates to synergy, and under which additional conditions it can guarantee—or be implied by—the emergence of synergistic human–AI performance.

To this end, we first propose a formalization of the notion of appropriate reliance, inspired by the metrics proposed in \cite{campagner2026metrics}, in our framework:
\begin{nameddef}[Appropriate Reliance (AR)]
\begin{align}
\text{AR1}) \quad \Pr(011) > \Pr(010), \quad \text{AR2}) \quad \Pr(101) > \Pr(100). \nonumber
\end{align}
\end{nameddef}
The two conditions, respectively, denote the lack of self-anchoring bias (also called egocentric discounting \cite{yaniv2000advice}) (AR1), as the probability to accept a correct AI support is higher than the probability of rejecting it, and the lack of automation bias \cite{parasuraman1997humans} (AR2), as the probability to discard an incorrect AI support is higher than the probability of accepting it.

Based on the assumption of appropriate reliance, we can prove several alternative characterizations of Synergy.

\begin{theorem}
\label{th:ar}
    AR, BA imply Synergy under one of the following:
   \begin{enumerate}
       \item $\Pr(100) = \Pr(010)$;
       \item $\Pr(101) = \Pr(011)$;
       \item $\Pr(100) > \Pr(010)$ and Algorithmic Advantage;
       \item $\Pr(010) > \Pr(100)$ and Human Advantage;
       \item $\Pr(011) > \Pr(101)$ and Human Advantage;
       \item $\Pr(101) > \Pr(011)$ and Algorithmic Advantage;
       \item $\Pr(100) > \Pr(010)$ and $\Pr(101) > \Pr(011)$;
       \item $\Pr(010) > \Pr(100)$ and $\Pr(011) > \Pr(101)$
   \end{enumerate}
   Furthermore, in results 3 and 5 (resp., 4 and 6) above, AR can be weakened to AR2 (resp., AR1).
   Under Agreement and one of conditions 1, 2, 7, or 8, Synergy implies AR. Furthermore, under condition 7, Synergy can relaxed to S2; under condition 8, Synergy can be relaxed to S1.
\end{theorem}
The previous result shows that, while in the literature Synergy and Appropriate Reliance are often related to each other, they are not equivalent unless other conditions are satisfied. Interestingly, the result, together with Proposition 3, illustrate conditions for establishing Human (resp., Algorithmic) Correction: ensuring lack of automation bias (resp., self-anchoring bias) and $\Pr(100) > \Pr(010)$ (resp., $\Pr(010) > \Pr(100)$), where this latter condition can be seen as a weak form of Human (resp., Algorithmic) Advantage. Nonetheless, understanding which algorithmical or socio-technical interventions can be implemented to ensure such a guarantee is not trivial, and it represents a relevant direction for future research.

\section{Cognitive and Metacognitive Correlates of Human Correction}

The notion of \emph{Human Correction} (HC), introduced in Section~4, captures a key aspect of the human contribution to Synergy: the ability of a human decision-maker to correctly override an algorithmic output when the latter is wrong. Formally, HC is satisfied when, conditional on a disagreement between human and algorithmic assessments, the final decision $F$ follows the human judgment $H$ more often when the AI is wrong, than when it is correct. In other words, HC expresses a systematic capability to detect and correct algorithmic errors.

Beyond its formal definition, this property has a precise cognitive interpretation. HC can be viewed as the behavioral manifestation of a second-order competence, namely \emph{metacognitive sensitivity} --- the capacity to monitor and evaluate one's own confidence and the reliability of external sources. In the context of human--AI interaction, metacognitive sensitivity supports an adaptive balance between reliance and autonomy: it allows the human agent to infer, from contextual cues and subjective uncertainty, when the algorithm is likely to be wrong and when it can be trusted. This self-monitoring faculty is pivotal to Synergy, empowering humans to provide informed interventions instead of arbitrary rejection.

In the literature on metacognition, such sensitivity is often quantified through \emph{Type~2 receiver operating characteristic} (Type~2 ROC) analysis, which assesses the correspondence between confidence judgments and actual correctness \cite{FlemingLau2014}. A high area under the Type~2 ROC curve indicates that confidence reliably tracks accuracy, implying that the agent can discriminate between correct and incorrect decisions. Translating this into our probabilistic framework, a high metacognitive sensitivity corresponds to a high conditional probability $\Pr(H = X \mid \text{high confidence})$ and, consequently, can increase the likelihood that the HC condition holds in practice, insofar as it enables selective overriding when contextual cues indicate that the algorithm may be wrong. However, we note that HC is a behavioral property of the interaction protocol and cannot be reduced to confidence discrimination alone.

This interpretation also connects HC to the notion of \emph{confidence calibration}, a classical measure of the alignment between subjective confidence and objective accuracy. Poor calibration can lead either to \emph{automation bias}, when overconfidence in the AI diminishes corrective intervention, or to \emph{self-anchoring bias} (also called \emph{algorithmic aversion}~\cite{Burton2020AlgorithmAversion} or \emph{conservatism bias} \cite{Cabitza2023}), when underconfidence prevents beneficial reliance. By contrast, well-calibrated metacognitive monitoring fosters appropriate reliance which, under certain conditions (as shown in Theorem 3), can lead to synergistic effects. 
Thus, while Algorithmic Correction (AC) depends on the statistical superiority of the model (\emph{Algorithmic Advantage}), Human Correction (HC) depends on the human's ability to track and regulate the reliability of both agents in interaction.

This perspective underscores that synergy is not solely a product of informational overlap, but also of the human’s metacognitive competence to exploit that complementarity~\cite{steyvers2022bayesian}.
This competence is part and parcel of the cognitive capability that has been also denoted as \emph{teaming intelligence}, that is the kind of intelligence and competence needed to effectively work with AI systems~\cite{johnson2019teaming}, and \emph{theory of machine}\cite{logg2019algorithm,dietvorst2015algorithm}, that is a practical understanding of how the AI system behaves, including its strengths, limitations, and typical failure modes (i.e., a form of Theory of Mind directed toward machines). These ``theories'', while not necessarily technically accurate, are pragmatically valuable for interpreting the system’s behavior, evaluating the reliability of its outputs, and determining when to rely on or disregard its suggestions~\cite{introzzi2024enhancing}.

Accordingly, the empirical detection of Synergy should include measures of metacognitive sensitivity (e.g., confidence discrimination, Type~2~ROC area, appropriate reliance score), human-AI agreement (e.g., influence index), and calibration (ECE, ECI), alongside standard accuracy metrics. Integrating these psychometric and behavioral constructs into the probabilistic framework developed in this paper provides a bridge between formal models of judgment combination and cognitive theories of self-monitoring, offering a richer account of when and why genuine Human--AI Synergy can emerge.

\section{Empirical illustration: clinical case studies}
\label{sec:empirical}

In this section we describe an empirical validation of our theoretical framework through seven real-world case studies, across diverse clinical domains, drawn from the human-AI interaction literature. While the following analysis is not designed as a controlled experimental validation, it provides an empirical testbed to examine whether the theoretical conditions derived in this paper correspond to observable patterns in real-world human–AI interaction data, thus providing an ecological validation of our theoretical results.

All the studies involved the interaction between expert clinicians and (simulated or real) AI systems, adopting a human-first interaction protocol by which humans were initially asked to provide a diagnosis without any external support; secondly, an AI-generated support was generated and provided to the humans; finally, based on the their previous judgment and the AI-support, a final decision was made by the humans. This interaction modality enables the collection of reliance pattern data, which were either public or provided to us by the authors of the studies, and hence allows to empirically verify the presence or absence of synergy, as well as of the other conditions described in this work. The characteristics of the studies are as follows:

\begin{itemize}
    \item Knee MRI Diagnosis (MRI): In this orthopedic study, 12 radiologists from IRCCS Ospedale Galeazzi Sant'Ambrogio (Milan, Italy) and other hospitals in Italy, had to classify 120 knee lesions using MRI scans selected from the MRNet dataset\footnote{\url{https://stanfordmlgroup.github.io/competitions/mrnet/}}, supported by a simulated AI system. 
    Further details can be found in~\cite{cabitza2023rams}.
    \item Cardiologic ECG Evaluation (ECG): In this cardiological study, 21 medical professionals from the University Hospital of Siena (Italy) had to classify heartbeat patterns from 20 ECGs, supported by a simulated AI system. These ECGs were selected from the ECG Wave-Maven repository\footnote{\url{https://ecg.bidmc.harvard.edu/maven/mavenmain.asp}}. 
    Further details can be found in~\cite{cabitza2023rams}.
    \item Spinal X-Ray Diagnosis (X-Rays): In this orthopedic study, 7 orthopedists from the Gaetano Pini-CTO Hospital (Milan, Italy) had to detect traumatic thoraco-lumbar fractures 12 X-ray images, supported by a real AI system. T
    Further information is available in~\cite{cabitza2022color}. 
    \item Gastrointestinal Endoscopy Assessment (Endoscopy (HP), Endoscopy (IL), Endoscopy (UCEIS)):  This study is a large-scale endoscopic evaluation in which 274 medical professionals from all over Europe were involved in three sub-studies. In the HP study, participants rated 30 cases of potential gastrointestinal bleeding (Hemorrhagic Potential) in short (15–20 second) endoscopic colonoscopy videos. In IL, participants rated 30 cases of potential inflammatory lesions in small bowel capsule endoscopy. In UCEIS, participants rated 30 cases of ulcerative colitis activity using proctosigmoidoscopy. In all, participants were supported by a simulated AI system. Further information is available in \cite{cabitza2025five} 
    \item Colorectal Endoscopy Diagnosis (Endoscopy (Colorectal)): In this study, 21 endoscopists had to diagnose 504 video clips of colorectal lesions, supported by a real-world AI tool. Further information is available in \cite{Reverberi2022}.
    
\end{itemize}

\subsection{Results}
The reliance patterns are summarized in Table \ref{tab:results}.

\begin{table*}[!htb]
    \centering
    \caption{Results of the considered case studies, in terms of counts of reliance patterns.}
    \label{tab:results}
    \tiny
    \begin{tabular}{c|ccccccc}
            & MRI & ECG & X-Rays & Endoscopy (HP) & Endoscopy (IL) & Endoscopy (UCEIS) & Endoscopy (Colorectal)\\\hline
        000 & 82  & 168 & 4   & 363  & 460  & 182  & 1126\\
        001 & 1   & 1   & 10  & 4    & 3    & 52   & 245\\
        010 & 238 & 102 & 9   & 649  & 350  & 345  & 251\\
        011 & 46  & 71  & 38  & 376  & 184  & 174  & 827\\
        100 & 21  & 12  & 5   & 75   & 61   & 141  & 540\\
        101 & 202 & 71  & 45  & 410  & 336  & 494  & 954\\
        110 & 11  & 0   & 16  & 53   & 10   & 15   & 173\\
        111 & 947 & 415 & 161 & 3219 & 3076 & 2933 & 6464\\\hline
        UJ & 0.76 & 0.59 & 0.79 & 0.73 & 0.78 & 0.83 & 0.77 \\
AI & 0.80 & 0.70 & 0.78 & 0.83 & 0.81 & 0.80 & 0.73 \\
SD & 0.77 & 0.66 & 0.88 & 0.78 & 0.80 & 0.84 & 0.80 \\
    \end{tabular}
    
\end{table*}

Synergy emerged in only three out of seven studies: X-Rays, Endoscopy (UCEIS), and Endoscopy (Colorectal). In the four settings where synergy failed (MRI, ECG, Endoscopy (HP), and Endoscopy (IL)), Algorithmic Advantage ($AI > UJ$) was satisfied. However, Human Correction (HC) failed in all these cases. Notably, while Algorithmic Correction (AC) held in every study, its presence alone was insufficient to guarantee synergy in the absence of HC. This failure suggests a ``dominance trap'': a strong Algorithmic Advantage can inhibit synergy if the human expert fails to identify the specific ``broken leg'' cases where the AI errs, thereby failing to provide the necessary Human Correction.

In regard to the synergistic cases, two of them illustrate similar theoretical pathways. In Endoscopy (UCEIS), synergy was achieved (approx. 1.5\% improvement) through the simultaneous fulfillment of Human Advantage, HC, AC, and BA, thus providing empirical validation for Theorem \ref{th:hcacasyn}.
 Interestingly, this occurred despite a lack of full Appropriate Reliance: while Automation Bias was not dominant, a significant self-anchoring Bias persisted. This provides a critical insight: AR is neither a sufficient nor necessary path to synergy; a high rate of HC can compensate for a lack of condition AR1, provided that the combination rule ``vetoes'' the most critical AI errors. In Endoscopy (Colorectal), a robust synergistic effect (approx. 3\%) was observed. Similarly to the case of Endoscopy (UCEIS), also here, BA, Human Advantage, HC, and AC all held simultaneously. Unlike the UCEIS study, however, this case also demonstrated full appropriate reliance, aligning with Case 7 of Theorem \ref{th:ar}. 

The X-rays study presents the most unique scenario. Although HC and AC were satisfied, the interaction violated both Agreement and BA. Despite this, a substantial 9\% accuracy gain was recorded. This suggests that the quality of human-AI interaction is not solely about the presence of (human or algorithmic) correction but rather about their ``magnitude''. In this case, which provides an illustration of Theorem \ref{th:synergyeq}, the \emph{strength} of both HC and AC was sufficiently large to overcome sporadic non-agreement. This implies that synergy is driven by the diagnostic resolution of the team: when HC and AC are strong, the human and AI outputs are effectively de-correlated on their respective errors, maximizing the information gain of the hybrid system.
In turn, this provides inspiration for a generalization of our results, which gives a sufficient and necessary condition for Synergy:
\begin{theorem}
\label{th:synergyeq}
    Synergy is equivalent to the following conditions:
    \begin{align}
        &\Pr(101) - \Pr(010) > \Pr(110) - \Pr(001), \nonumber\\
        &\Pr(011) - \Pr(100) > \Pr(110) - \Pr(001)\nonumber
    \end{align}
\end{theorem}
Theorem~\ref{th:synergyeq} shows that synergy arises from a trade-off. In both inequalities, the left-hand side can be interpreted as the \emph{magnitude of correction}, quantifying how strongly Human and Algorithmic Correction operate. By contrast, the right-hand side captures an \emph{aversion penalty}. This penalty is $0$ under agreement and becomes negative under Beneficial Aversion. Hence, synergy can emerge only when the combination function integrates the partial judgments $H$ and $A$ with context-dependent information about the situation $S$ in such a way that the magnitude of correction outweighs the penalty associated with non-agreement (if any). Thus, while imposing structural constraints, such as agreement, on the combination rule may provide a simplified pathway to synergy, this latter can nonetheless be reached even when such constraints fail, provided that the interaction policy implemented by the combination rule is able to enforce the trade-off in Theorem \ref{th:synergyeq}.

\section{Discussion and Implications for the Design of AI-based Decision Support}


Our theoretical and empirical findings support a central conclusion: synergy is not primarily determined by the standalone accuracy of human or AI agents, but is instead an emergent property of the interaction policy that governs how their judgments are combined. This perspective shifts the focus of system design from improving individual models in isolation to engineering interaction mechanisms that enable effective collaboration between humans and AI systems.

A key implication of our results is that increasing the performance of an AI system alone is not sufficient to guarantee improved system-level outcomes. In fact, the prevalence of the “dominance trap” in several of the clinical case studies, suggests that strong AI performance can inadvertently suppress human involvement in the decision process, in particular when the human user \emph{owns} the final process of decision combination. When users become overly reliant on the AI, opportunities for human correction diminish, and the system as a whole may fail to exploit complementary strengths. This highlights the importance of designing interaction protocols that preserve an active and meaningful role for human judgment, even when AI performance is high. A practical line of intervention would consist of avoiding creating black-box systems that merely provide a final label. Instead, AI tools should be designed to enhance users’ understanding of their outputs. However, this does not entail aligning AI system design uncritically with XAI research, which maintains that it is important for AI systems to ``explain'' what they do. A growing body of literature (e.g.,~\cite{cabitza2024explanations,ehsan2024explainability,ghassemi2021false}) has already warned about the risks of machine explanations increasing deference and automation bias, precisely because of their persuasive power. In this sense, our framework suggests treating explanations as reliability metadata that inform users at the level of individual classifications and local predictions~\cite{peracchio2025relai}, or as alternative hypotheses~\cite{miller2023explainable,blom2025improving} to consider in order to gain a better understanding of the matter at hand. In doing so, AI systems would promote users’ metacognitive sensitivity \cite{lee2025metacognitive}—for instance, by highlighting areas of high uncertainty or providing cues that enable humans to effectively perform Human Correction.

Furthermore, while we note that enforcing the agreement condition is trivial from a systems-design perspective\footnote{Indeed, agreement can achieved by merely activating the combination rule only when a disagreement arises.}, the evidence that Synergy can emerge even in the absence of such conditions suggests that the goal of a hybrid system should be ``functional de-correlation'', as this latter can enforce even larger gains in performance than merely relying on interaction policy that are solely relegated to cases of disagreement. This insight has different implications depending on whether the combination rule is realized by human or, rather, through computational means. In the first case, human training can be easily seen to be a fundamental dimension for enabling synergy: users of AI-based decision support systems should develop a properly calibrated theory of machine. Training programs that expose users to examples of AI errors, or that simulate disagreement scenarios, can help improve calibration and support more effective reliance behavior. Such training can enhance the ability of users to detect when the AI is likely to be correct and when it should be overridden. By contrast, in the second case, our results imply that AI systems should be optimized not to simulate human reasoning—whatever this may mean—but rather to develop a distinct perspective on the phenomena of interest \cite{wilder2020learning}. In this sense the conditions of Human and Algorithmic Correction, as well as the margin conditions given in Theorem \ref{th:synergyeq}, provide natural learning objectives to be used to  to tailor a policy implementing the combination rule. More in general, and in both cases, our results also imply that designers should focus on identifying and supporting the human’s ability to act as a diagnostic filter, directing attention to cases in which the AI is most likely to fail (the ``broken leg'' cases) and protecting expertise from skill atrophy.
This latter concern suggests designing and implementing human–AI interaction protocols aimed at ``slowing down'' users and making interaction with AI systems more effortful~\cite{booch2021thinking}—so-called inhibition protocols, that are, protocols that deliberately introduce a form of \emph{cognitive friction}~\cite{cabitza2023let}, such as requiring users to provide their judgment before being exposed to the machine’s advice~\cite{cabitza2025five}, or having the AI system abstain from providing any advice at all when the risk of broken leg cases is too high \cite{mozannar2023who}, with goal of making engagement with AI a more deliberative and reflective (rather than automatic) process. 

At a broader level, governance and deployment strategies should take into account not only the accuracy of individual components but also the quality of their interaction. Monitoring system performance should include metrics derived from reliance patterns, disagreement rates, and correction frequencies. These indicators provide insight into whether the system operates in a regime conducive to synergy, and can reveal whether interaction policies are functioning as intended. Over time, such monitoring can support continuous improvement of both the AI system and the surrounding interaction protocols.

In summary, achieving synergy requires a coordinated approach that integrates model development, interface design, interaction protocols, and user training. Rather than focusing solely on maximizing standalone performance, system designers should aim to create environments in which human and AI agents can effectively complement each other. This involves promoting structured interaction, designing principled disagreement-resolution policies, supporting metacognitive awareness, and encouraging functional differences that allow each agent to correct the other. Under these conditions, synergy emerges not as an accidental outcome but as a design-enabled property of the human-AI system.

\section{Conclusion}

This paper has developed a probabilistic framework to formalize the elusive notion of human-AI synergy. Our analysis demonstrates that while Algorithmic Advantage and Human Augmentation are frequent benchmarks in the literature, true Synergy is a more rigorous condition, mathematically tied to the existence of effective Human Correction (HC) and Algorithmic Correction (AC) mechanisms. By grounding these concepts in the formal calculus of reliance patterns, we have provided a principled basis to move beyond simple accuracy metrics toward a deeper understanding of hybrid decisional systems, as well as highlighted significant implications for the design of hybrid human-AI systems.

Our empirical investigation across seven clinical datasets confirms that Synergy is not a guaranteed outcome of AI integration. The observed ``dominance trap'' in high-performing AI settings and the compensatory role of HC and AC highlight that the path to Synergy is mediated by the human's metacognitive sensitivity and the team's ability to functionally de-correlate errors. These findings suggest that the future of AI design should prioritize the calibration of the human-AI interaction over the mere maximization of model performance.

In conclusion, the pursuit of Synergy, although more challenging than plain automation, is shown to be a mathematically sound and empirically attainable goal. The formal conditions identified here provide researchers and practitioners with a roadmap for designing and evaluating hybrid systems that do not merely supplement human judgment but transcend the limitations of both human and artificial intelligence, to realize the potential of a \emph{hybrid intelligence}.

\bibliographystyle{IEEEtran}
\bibliography{biblio}

\end{document}


\newcommand{\uj}{\text{UJ}}
\newcommand{\ai}{\text{AI}}
\newcommand{\sd}{\text{SD}}
\markboth{IEEE Transactions on Human-Machine Systems,
Vol. X, No. X, March 2026}
{A. Campagner
\MakeLowercase{\textit{(et al.)}:
When Do Humans and Machines Truly Complement Each Other?}}

\title{When Do Humans and Machines Truly Complement Each Other? Formal Conditions and Empirical Evidence for Human-AI Synergy}

\author{Andrea Campagner, Federico Cabitza, Vincenzo Crupi \thanks{A. Campagner is with Department of Informatics, Systems and Communication, University of Milano-Bicocca, Milan, Italy and IRCCS Ospedale Galeazzi Sant'Ambrogio, Milan, Italy} \thanks{F. Cabitza is with Department of Informatics, Systems and Communication, University of Milano-Bicocca, Milan, Italy and Digital Health \& Wellbeing Center, Fondazione Bruno Kessler (FBK), Trento, Italy} \thanks{V. Crupi is with Department of Philosophy and Education Sciences, University of Turin, Torino, Italy}}




\maketitle
\begin{abstract}
Can humans and artificial intelligence cooperate such that their combination yields greater accuracy than either agent alone? This result, termed \emph{synergy} or \emph{complementarity}, is notoriously difficult to achieve, even when both agents possess significant independent accuracy. In this paper, we present a formal framework to model human-AI interaction through the lens of reliance patterns, which characterize the joint correctness structure of human and algorithmic decisions. We derive mathematically necessary and sufficient conditions for synergy, demonstrating that its emergence depends on the ability of either agent to identify and correct the errors of the other, what we call Human and Algorithmic Correction. Beyond its theoretical contribution, the proposed framework provides a foundation for the design and evaluation of knowledge-based decision support systems, where interaction policies must dynamically integrate heterogeneous sources of expertise. We demonstrate this through the analysis of seven clinical case studies. Our empirical results reveal a ``dominance trap,'' where a strong algorithmic advantage can paradoxically inhibit synergy by eroding human corrective intervention. We also show that synergy can persist even in the presence of cognitive biases, such as automation bias, provided there is functional de-correlation between human and AI errors. These results provide a formal foundation for key concepts in the human-AI systems literature and offer actionable insights for designing interaction-centric AI systems that foster complementarity.
\end{abstract}

\begin{IEEEkeywords}
   Human-AI Synergy, Human-AI Complementarity, Human-AI Interaction, Reliance Patterns, Metacognition, Dominance Trap
\end{IEEEkeywords}

\setcounter{theorem}{0}
\setcounter{proposition}{0}
\section{Proofs and Additional Results}


\begin{proposition}
    Assume the default setting. If $c$ satisfies \text{Agr} then $\Pr(F = X) > 0.25$.
\end{proposition}
\begin{proof}
$$\Pr(F = X) \geq \Pr(H = A = X) \quad[Agr].$$
Additionally, by CI, it holds that:
    \begin{align}
       &\Pr(H = A = X) =\nonumber\\ 
       &\Pr(H=X|X=x)\Pr(A=X|X=x)\Pr(X=x) +\nonumber\\
       &\Pr(H=X|X=\neg x)\Pr(A=X|X= \neg x)\Pr(X=\neg x)\nonumber
    \end{align}
Finally, by HR and AR:
        $$0.25\Pr(X=x) + 0.25\Pr(X=\neg x) \geq 0.25.$$
\end{proof}

\begin{proposition1}
    Assume the default setting. If $c$ is either AA or FA, then $\Pr(F = X) > 0.5$.
    
    Furthermore, let $p = \Pr(X=x)$. If $c$ is DS, then $\Pr(F=X) > 0.75 - 0.5p$: in particular, if $p < 0.5$, then  $\Pr(F=X) > 0.5$. If $c$ is DA, then $\Pr(F=X) > 0.5p + 0.25$: if $p > 0.5$, then $\Pr(F=X) > 0.5$.
\end{proposition1}
\begin{proof}
    For AA:
    \begin{align}
    &\Pr(F = X) = \Pr(H = X) \quad[AA] \nonumber\\
    &\Pr(F = X) > 0.5 \quad [HR]. \nonumber
    \end{align}
    The argument for FA is similar, but replacing HR for AR.

    For DS, by the law of total probability:
    \begin{align}
    \Pr(F = X) &= \Pr(F=f|X=x)\Pr(X=x) \nonumber\\
    &+ \Pr(F=\neg f | X = \neg x)\Pr(X=\neg x) \nonumber
    \end{align}
    The first term of the summation is $\Pr(A=a \wedge H=h|X=x)\Pr(X=x)$, which, by CI, HR and AR, is larger than $0.25\Pr(X=x)$. The second term is $1 - \Pr(F=f| X = \neg x)\Pr(X=\neg x) = 1 - \Pr(A=a \wedge H=h|X=\neg x)\Pr(X=\neg x)$ which, by the previous argument, is larger than $0.75\Pr(X=\neg x)$. Thus, denoting $p=\Pr(X=x)$:
    \begin{align}
    \Pr(F = X) &\geq 0.75(1-p) + 0.25p = 0.75 - 0.5p. \nonumber
    \end{align}

    For DA, we employ the same decomposition, by the law of total probability, as used for DS. In this case, by CI, HR and AR, the first term of the summation is $\Pr(A=a \vee H=h|X=x)\Pr(X=x)$, where $\Pr(A=a \vee H=h|X=x) = \Pr(A=a|X=x) + \Pr(H=h|X=x) - \Pr(A=a \wedge H=h|X=x) > 0.75$. The second term, by contrast, is $\Pr(A=\neg a \wedge H=\neg h|X=\neg x)\Pr(X=\neg x)$ which, by the previous argument, is larger than $0.25(1 - \Pr(X=x))$. Letting $p=\Pr(X=x)$ as before:
    \begin{equation}
        \Pr(F=X) > 0.75p + 0.25(1-p) = 0.5p + 0.25.
    \end{equation}
    
\end{proof}

\begin{theorem}
\label{th:hcacasyn}
     Assuming Human and Algorithmic Correction, the Agreement property becomes a sufficient condition for Synergy. Similarly, assuming Agreement, Human and Algorithmic Correction are sufficient for Synergy. Conversely, Synergy and Agreement imply Human and Algorithmic Correction.
\end{theorem}
\begin{proof}
For S1:
\begin{align}
    &Pr(101) \geq  Pr(010) \quad[HC] \nonumber\\
    &Pr(101) + Pr(001) > Pr(010) + Pr(110) \quad[Agr] \nonumber
\end{align}
Similarly, for S2:
\begin{align}  
    &P(011) \geq P(100) \quad[AC]\nonumber\\
    &P(011) + Pr(001) \geq P(100) + Pr(110) \quad [Agr]\nonumber
\end{align}

Conversely:
\begin{align}
    &\Pr(101) + \Pr(001) > \Pr(010) + \Pr(110) \quad [S1] \nonumber\\
    &\Pr(101) > Pr(010) \quad [Agr] \nonumber\\
    &\Pr(011) + Pr(001) \geq P(100) + Pr(110) \quad [S2] \nonumber\\
    &\Pr(011) > Pr(100) \quad [Agr] \nonumber
\end{align}
\end{proof}

\begin{theorem}
    Assume Human Correction and Algorithmic Correction. Then Beneficial Aversion is sufficient for Synergy. The converse part of Theorem \ref{th:hcacasyn}, however, does not hold.
\end{theorem}
\begin{proof}
    The first part of the result follows directly from the proof of Theorem \ref{th:hcacasyn}. Indeed by replacing [Agr] for [BA] in the proofs that HC implies S1 and AC implies S2, we obtain the result.
    
    For the second part, it is easy to observe that [BA] cannot be used in the same way as [Agr] to eliminate the factors $\Pr(001), \Pr(110)$ from the expressions appearing in the proof. Indeed, we show by counterexample cases in which the converse results does not hold. Set $\Pr(101) = 0.3, \Pr(010) = 0.1, \Pr(001) = 0.1, \Pr(110) = 0.2$. Then S1  and HC holds, but BA does not hold. An equivalent result follows similarly also for S2 and AC, by setting $\Pr(011) = 0.3, \Pr(100) = 0.1, \Pr(001) = 0.1, \Pr(110) = 0.2$. The result follows.
\end{proof}

\begin{proposition}[HC $\Leftrightarrow$ Synergy under Algorithmic Advantage and Agreement]
\label{prop:hcsmpa}
Assume Algorithmic Advantage and Agreement. Then Human Correction is necessary and sufficient for Synergy. It is sufficient also if Agreement is replaced by Beneficial Aversion, or if Algorithmic Advantage is replaced by Meehl's Postulate.
\end{proposition}
\begin{proof}
For sufficiency:
\begin{align}
    &\Pr(101) \geq  \Pr(010) \quad [HC] \nonumber\\
    &\Pr(101) + \Pr(001) > \Pr(010) + \Pr(110) \quad [Agr] \nonumber\\
    &\Pr(F=X) > \Pr(A=X) > \Pr(H = X) \quad [A-Adv] \nonumber
\end{align}
In the second step of the proof above, it is easy to see that [Agr] can be replaced by [BA]. Similarly, since Meehl's Postulate is stronger than Algorithmic Advantage, in the third step of the proof above, [A-Adv] can be replaced by [MP]. 

For necessity, by Synergy and Algorithmic Advantage we have $Pr(F=X) > Pr(\ai=X) > Pr(H=X)$. In particular, the first inequality simplifies to S1 ($Pr(101) + Pr(001) > Pr(010) + Pr(110)$), from which, by Agreement, we obtain $Pr(101) > Pr(010)$.
\end{proof}

\begin{proposition}[AC $\Leftrightarrow$ Synergy under Human Advantage and Agreement]
Assume Human Advantage and Agreement. Then Human Correction is necessary and sufficient for Synergy. It is sufficient also if Agreement is replaced by Beneficial Aversion.
\end{proposition}
\begin{proof}
The proof is equivalent to the proof of Prop. \ref{prop:hcsmpa}.
\end{proof}

\begin{theorem}
   AR, BA imply Synergy if one of the following holds:
   \begin{enumerate}
       \item $\Pr(100) = \Pr(010)$;
       \item $\Pr(101) = \Pr(011)$;
       \item $\Pr(100) > \Pr(010)$ and Algorithmic Advantage;
       \item $\Pr(010) > \Pr(100)$ and Human Advantage;
       \item $\Pr(011) > \Pr(101)$ and Human Advantage;
       \item $\Pr(101) > \Pr(011)$ and Algorithmic Advantage;
       \item $\Pr(100) > \Pr(010)$ and $\Pr(101) > \Pr(011)$;
       \item $\Pr(010) > \Pr(100)$ and $\Pr(011) > \Pr(101)$
   \end{enumerate}
   Furthermore, in results 3 and 5 (resp., 4 and 6) above, AR can be weakened to AR2 (resp., AR1).
   Under Agreement and either condition 1 or condition 2, or condition 7, or condition 8, Synergy implies AR. Furthermore, under condition 7, Synergy can relaxed to S2; under condition 8, Synergy can be relaxed to S1.
\end{theorem}
\begin{proof}
We prove the result under the assumption that $\Pr(100) = \Pr(010)$ (i.e., case 1). Case 2, in which $\Pr(101) = \Pr(011)$, is equivalent. For S1:
\begin{align}
    &\Pr(101) > \Pr(100) \quad [AR2]\nonumber\\
    &\Pr(101)+\Pr(001) > \Pr(100)+\Pr(110) \quad [BA]\nonumber\\
    &\Pr(101)+\Pr(001) > \Pr(010)+\Pr(110) \nonumber
\end{align}
Similarly, for S2:
\begin{align}
    &\Pr(011) > \Pr(010) \quad [AR1]\nonumber\\
    &\Pr(011)+\Pr(001) > \Pr(010)+\Pr(110) \quad [BA]\nonumber\\
    &\Pr(011)+\Pr(001) > \Pr(100)+\Pr(110)\nonumber
\end{align}

For case 3, assuming AR, XA, $\Pr(100) > \Pr(010)$, we obtain S1:
\begin{align}
    &\Pr(101) > \Pr(100) \quad [AR2]\nonumber\\
    &\Pr(101)+\Pr(001) > \Pr(100)+\Pr(110) \quad [BA]\nonumber\\
    &\Pr(101)+\Pr(001) > \Pr(010)+\Pr(110) \nonumber
\end{align}
By Algorithmic Advantage, we obtain $\Pr(F=X) > \Pr(A=X) > \Pr(H=X)$. Note that only AR2 has been employed in the proof.

For case 4, assuming AR, XA, $\Pr(010) > \Pr(100)$, we obtain S2:
\begin{align}
    &\Pr(011) > \Pr(010) \quad [AR1]\nonumber\\
    &\Pr(011)+\Pr(001) > \Pr(010)+\Pr(110) \quad [BA]\nonumber\\
    &\Pr(011)+\Pr(001) > \Pr(100)+\Pr(110) \nonumber
\end{align}
By Human Advantage, we obtain $\Pr(F=X) > \Pr(H=X) > \Pr(A=X)$. Note that only AR1 has been employed in the proof.

For case 5, assuming AR, XA, $\Pr(011) > \Pr(101)$, we obtain S2:
\begin{align}
    &\Pr(101) > \Pr(100) \quad [AR2]\nonumber\\
    &\Pr(101)+\Pr(001) > \Pr(100)+\Pr(110) \quad [BA]\nonumber\\
    &\Pr(011)+\Pr(001) > \Pr(100)+\Pr(110) \nonumber
\end{align}
By Human Advantage, we obtain $\Pr(F=X) > \Pr(H=X) > \Pr(A=X)$. Note that only AR2 has been employed in the proof.

For case 6, assuming AR, XA, $\Pr(101) > \Pr(011)$, we obtain S1:
\begin{align}
    &\Pr(011) > \Pr(010) \quad [AR1]\nonumber\\
    &\Pr(011)+\Pr(001) > \Pr(010)+\Pr(110) \quad [BA]\nonumber\\
    &\Pr(101)+\Pr(001) > \Pr(010)+\Pr(110) \nonumber
\end{align}
By Algorithmic Advantage, we obtain $\Pr(F=X) > \Pr(A=X) > \Pr(H=X)$. Note that only AR1 has been employed in the proof.

Case 7 follows by combining cases 3 and 6, while case 8 follows by combining cases 4 and 5.

For the equivalence between Synergy and AR under Agreement, consider case 1. By Theorem \ref{th:hcacasyn}, Synergy and Agreement imply HC and AC. Hence, for AR2:
\begin{align}
&\Pr(101) > \Pr(010) \quad [HC]\nonumber\\
&\Pr(101) > \Pr(100)\nonumber
\end{align}
Similarly, for AR1:
\begin{align}
&\Pr(011) > \Pr(100) \quad [AC]\nonumber\\
&\Pr(011) > \Pr(010)\nonumber
\end{align}
Case 2 similarly follows.

Consider now the condition in case 7. Then, S2 and Agreement imply AC, which, under the assumption that $\Pr(100) > \Pr(010)$, implies AR1. Then, from AC and $\Pr(101) > \Pr(011)$, AR2 follows. The result for the condition in case 8 similarly follows.
\end{proof}

\begin{theorem}
    Synergy is equivalent to the following set of conditions:
    \begin{align}
        &\Pr(101) - \Pr(010) > \Pr(110) - \Pr(001), \nonumber\\
        &\Pr(011) - \Pr(100) > \Pr(110) - \Pr(001)\nonumber
    \end{align}
\end{theorem}
\begin{proof}
    The result directly follows from the definition of Synergy. In particular, the first conditions is equivalent to S1, while the second is equivalent to S2.
\end{proof}